\documentclass[journal,onecolumn,12pt,draftclsnofoot]{IEEEtran}

\usepackage{amsmath}
\usepackage{amsfonts}
\usepackage{amssymb}
\usepackage{amsthm}
\usepackage{color}
\usepackage{array}
\usepackage{cite}
\usepackage{enumerate}
\usepackage{graphicx}
\usepackage[hang]{subfigure}
\usepackage{epstopdf}
\usepackage{epsfig}
\usepackage{footmisc}
\usepackage{amsbsy}
\usepackage{bm}
\usepackage{cases}
\usepackage{multirow}
\usepackage{algorithmic}
\usepackage{comment}
\usepackage{times}
\usepackage{paralist}
\usepackage[ruled]{algorithm2e}
\usepackage{setspace}

\newcommand{\bfx}{{\mathbf x}}

\newcommand{\bfs}{{\mathbf s}}

\newcommand{\rmx}{\mathrm{x}}

\newcommand{\bmx}{{\bm x}}
\newcommand{\bms}{{\bm s}}

\newcommand{\bmr}{{\bm r}}

\newcommand{\bfU}{\mathbf{U}}

\newcommand{\bfV}{\mathbf{V}}
\newcommand{\bfW}{\mathbf{W}}

\newcommand{\bmtheta}{\bm{\theta}}

\newcommand{\bmalpha}{\bm{\alpha}}

\newcommand{\set}[1]{\ensuremath{\mathcal #1}}

\newcommand{\separator}{
  \begin{center}
    \rule{\columnwidth}{0.3mm}
  \end{center}
}

\def\eg{{\it e.g.}}
\def\ie{{\it i.e.}}

\newcommand{\beq}{\begin{eqnarray*}}
\newcommand{\eeq}{\end{eqnarray*}}
\newcommand{\beqn}{\begin{eqnarray}}
\newcommand{\eeqn}{\end{eqnarray}}
\newcommand{\bemn}{\begin{multiline}}
\newcommand{\eemn}{\end{multiline}}

\newcommand{\grad}[1]{\nabla #1}

\begin{document}
\title{Training Dynamic Exponential Family Models \\ with Causal and Lateral Dependencies \\ for Generalized Neuromorphic Computing}

\author{Hyeryung~Jang and Osvaldo~Simeone\thanks{H. Jang and O. Simeone are with the Department of Informatics, King's College London, London, United Kingdom (emails: hyeryung.jang@kcl.ac.uk, osvaldo.simeone@kcl.ac.uk).}
\thanks{This work was supported by the European Research Council (ERC) under the European Union's Horizon 2020 research and innovation programme (grant agreement No. 725731).}
}

% make the title area
\maketitle
\begin{abstract}
Neuromorphic hardware platforms, such as Intel's Loihi chip, support the implementation of Spiking Neural Networks (SNNs) as an energy-efficient alternative to Artificial Neural Networks (ANNs). SNNs are networks of neurons with internal analogue dynamics that communicate by means of binary time series. In this work, a probabilistic model is introduced for a generalized set-up in which the synaptic time series can take values in an arbitrary alphabet and are characterized by both causal and instantaneous statistical dependencies. The model, which can be considered as an extension of exponential family harmoniums to time series, is introduced by means of a hybrid directed-undirected graphical representation. Furthermore, distributed learning rules are derived for Maximum Likelihood and Bayesian criteria under the assumption of fully observed time series in the training set.
\end{abstract}


\section{Introduction} \label{sec:intro}

The current dominant computing framework for supervised learning applications is given by feed-forward multi-layer Artificial Neural Networks (ANNs). These systems process real numbers through a cascade of non-linearities applied successively over multiple layers. It is well-understood that training and running ANNs for inference generally require a significant amount of resources in terms of space and time (see, e.g., \cite{davies2018loihi}). Neuromorphic computing, currently backed by recent major projects by IBM, Qualcomm, and Intel, offers a fundamental paradigm shift that takes the trend towards distributed computing initiated by ANNs to its natural extreme by borrowing insights from computational neuroscience. A neuromorphic chip consists of a network of spiking neurons with internal temporal analogue dynamics and digital spike-based synaptic communications. Current hardware implementations confirm drastic power reductions by orders of magnitude with respect to ANNs \cite{diamond2016comparing, schuman2017survey}.

Models typically used to train Spiking Neural Networks (SNNs) are deterministic, and learning rules borrow tools and ideas from the design of ANNs. Examples include the standard leaky integrate-and-fire model and its variants, with associated learning rules that approximate backpropagation \cite{lee2016training, wu2018spatio,bohte2000spikeprop}. Probabilistic models for SNNs are more conventionally adopted in computational neuroscience and offer a variety of potential advantages, including flexibility and availability of principled learning criteria \cite{pillow2005prediction, koller2009pgm}. Nevertheless, they pose technical challenges in the design of training and inference algorithms that have only partially been addressed \cite{bagheri18:snn_first, gardner2016supervised, pfister2006optimal, rezende14:vi_snn}.

In this work, we study the problem of training a probabilistic model for correlated time series. The framework generalizes probabilistic models for SNNs \cite{hinton00:sBM,rezende14:vi_snn} by allowing for arbitrary alphabets -- not restricted to binary -- and by enabling individual time series -- or neuron signals -- not only to affect each other causally over time but also to have instantaneous correlations. It is noted that both general discrete alphabets and lateral dependencies are in principle implementable on existing digital neuromorphic chips \cite{davies2018loihi}. We derive distributed learning rules for Maximum Likelihood (ML) and for Bayesian criteria based on synaptic sampling \cite{neal11:mcmc,welling11:langevin,patterson13:langevin,sato14:langevin,kappel15:synaptic}, and detail the corresponding communication requirements. Applications of the model to supervised learning via classification are also discussed.

\begin{figure}[t]
  \hspace{0.3cm}
  \subfigure[]
  {
	\centering
	\includegraphics[height=0.2\columnwidth]{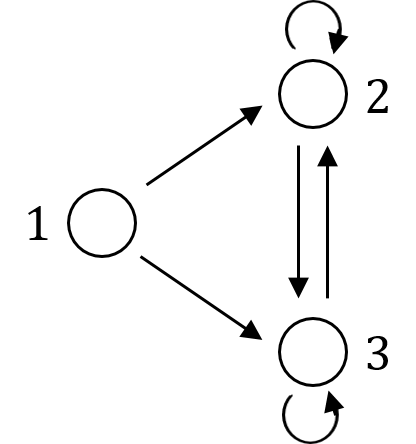}
    \label{fig:ex_causal_g}
  } 
  \hspace{0.2cm}
  \subfigure[]
  {
  \centering
    \includegraphics[height=0.2\columnwidth]{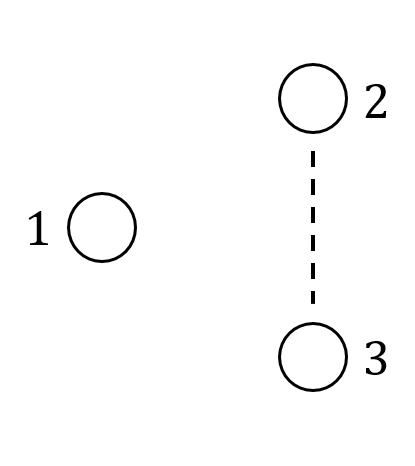} 
 	\label{fig:ex_lateral_g}
  }
  \hspace{0.2cm}
  \subfigure[]
  {
	\centering
\includegraphics[height=0.2\columnwidth]{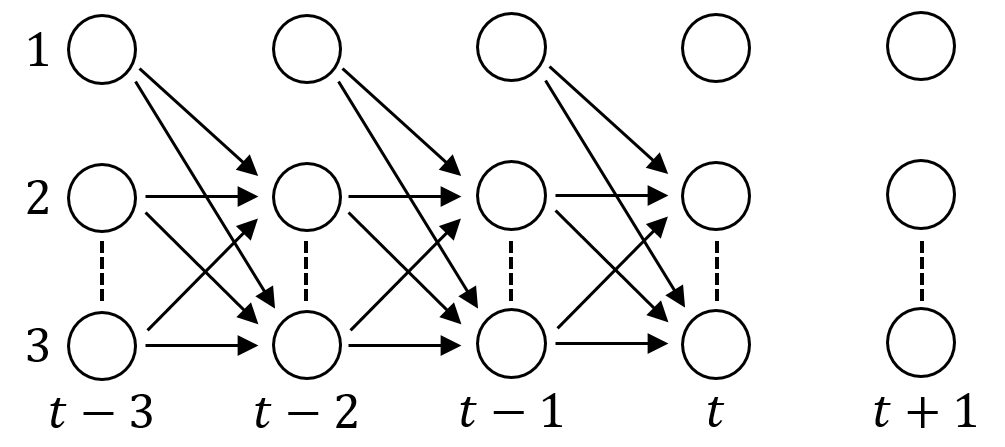}
  	\label{fig:ex_time_g}
  }
%  \hspace{-0.1cm}
\hfill
  \caption{Hybrid directed and undirected graphical representation of a dynamic exponential family model: (a) causal graph $\set{G}_{\set{P}}$ representing directed causal relationships; (b) lateral graph $\set{G}_{\set{L}}$ encoding instantaneous correlations; and (c) time-expanded graph assuming a memory for the causal connections equal to $\tau = 1$.}
  \label{fig:ex_model}
  %\vspace{-0.3cm}
\end{figure}

The proposed model can be considered as an extension of exponential family harmoniums \cite{welling05:harmoniums} from vector-valued signals to time series. We refer to the model as {\em dynamic exponential family}, which is specified by a hybrid directed and undirected graphical representation (see Fig.~\ref{fig:ex_model}). We specifically focus here on the case of fully observed time series, which is of independent interest and also serves as building block for the more complex set-up with latent variables, to be considered in follow-up work.

The rest of this paper is organized as follows. In Sec.~\ref{sec:model}, we describe the dynamic exponential family model. Under this model, we then derive distributed learning rules for ML and Bayesian criteria in Sec.~\ref{sec:learning}. Finally, Sec.~\ref{sec:numerical} presents numerical results for a multi-task supervised learning problem tackled via a two-layer SNN. 

\section{Dynamic Exponential Family Model} \label{sec:model}

In this section, we describe the proposed probabilistic model as an extension of exponential family harmoniums \cite{welling05:harmoniums} to time series. 

\smallskip
\noindent {\bf Hybrid directed-undirected representation.} 
We consider a probabilistic model for dependent time series that captures both causal and lateral, or instantaneous, correlations. To describe both causal and lateral dependencies, we introduce a directed graph $\set{G}_{\set{P}} = (\set{V}, \set{E}_{\set{P}})$ and an undirected graph $\set{G}_{\set{L}} = (\set{V}, \set{E}_{\set{L}})$, respectively, where $\set{V} = \{1, \ldots, N_x\}$ is the set of time sequences of interest. Having in mind the application to SNNs discussed above, we will also refer to the vertices $\set{V}$ as neurons or units. The edge set $\set{E}_{\set{P}} \subseteq \set{V}^2$ represents the {\em causal} connection between time series and corresponding neurons, and hence we refer to $\set{G}_{\set{P}}$ as the causal graph. Note that the presence of self-loops, \ie, edges of the form $(i,i)$ connecting a unit $i$ to itself, or of more general loops involving multiple nodes, is allowed in the causal graph, and it indicates a recurrent behavior for the time series. In contrast, the edge set $\set{E}_{\set{L}} \subseteq \set{V}^2$ encodes instantaneous correlations or {\em lateral} connections, and hence the graph $\set{G}_{\set{L}}$ is referred to as the lateral graph. Examples are shown in Fig.~\ref{fig:ex_causal_g} and Fig.~\ref{fig:ex_lateral_g} for a system with $N_x = 3$ time series and thus $N_x = 3$ nodes.

In the causal graph $\set{G}_{\set{P}},$ we denote by $\set{P}_i := \{j: (j,i) \in \set{E}_{\set{P}} \}$ the subset of units that has a causal connection to unit $i$. Note that the set $\set{P}_i$ includes also unit $i$ if there is a self-loop for unit $i$ and that a causal connection $(j,i) \in \set{E}_{\set{P}}$ from unit $j$ to unit $i$ does not imply the inclusion of the edge $(i,j)$ in $\set{E}_{\set{P}}$ in general. For the lateral graph $\set{G}_{\set{L}}$, we denote by $\set{L}_i := \{j: (i,j) \in \set{E}_{\set{L}} \}$ the subset of units that has a lateral connection with unit $i$. Edges $(i,j)$ and $(j,i)$ are equivalent in graph $\set{E}_{\set{L}}$. Moreover, we denote by $\set{C}_i$ the subset of units that are reachable from unit $i$ in lateral graph $\set{G}_{\set{L}}$ in the sense that there exists a path in $\set{G}_{\set{L}}$ between units $i$ and any unit $m \in \set{C}_i$, where a path is a sequence of edges in $\set{E}_{\set{L}}$ that share a vertex. We generally have the inclusion $\set{L}_i \subseteq \set{C}_i$; and that, for any $j \in \set{L}_i$, we have the set equality $\set{C}_i \cup \{i\} = \set{C}_j \cup \{j\}$, since two units $j$ and $i$ are laterally connected to each other. In example of Fig.~\ref{fig:ex_lateral_g}, we have the subsets $\set{C}_1 = \emptyset$, $\set{C}_2 = \{3\}$, and $\set{C}_3 = \{2\}$.

\smallskip
\noindent {\bf Probabilistic model.} 
Each unit $i \in \set{V}$ is associated with a random sequence $\rmx_{i,t}$ for $t=1,2,\ldots$, taking values in either a continuous or discrete space $\set{X}_i$. Given causal and lateral connections defined by graphs $\set{G}_{\set{P}}$ and $\set{G}_{\set{L}}$, the joint distribution of the random process $\bfx_t= (\rmx_{1,t}, \ldots, \rmx_{N_x,t})$ for $t=1,2,\ldots$ factorizes according to the chain rule 
\begin{align} \label{eq:DEF-joint-time}
p_\Theta(\bmx^T) = \prod_{t=1}^T p_\Theta(\bmx_t|\bmx^{t-1}),
\end{align}
where we denote as $\bfx^t=(\bfx_1,\ldots,\bfx_t)$ the overall random process from time $1$ to $t$, and as $\Theta$ the set of model parameters. In \eqref{eq:DEF-joint-time}, we have implicitly conditioned on initial value $\bmx_0$, which is assumed to be fixed. Furthermore, the conditional probability $p_\Theta(\bmx_t|\bmx^{t-1})$ follows a distribution in the exponential family with $N_{a}$-dimensional sufficient statistics $\bms_i(x_{i}) := [s_{i,1}(x_{i}), \ldots, s_{i,N_a}(x_{i})]^\top$ for every unit $i$ and a quadratic energy function. In the practically relevant case for neuromorphic computing where the alphabets $\set{X}_i$ are discrete and finite, \ie, $\set{X}_i = \{0,1,\ldots, C-1\}$ for some integer $C$, the sufficient statistics $\bms_i$ are defined as the one-hot representation, i.e., $\bms_i(x_i) = [1_{\{x_i = 1\}}, \ldots, 1_{\{x_i = C-1\}}]^\top$, where $1_{E}$ is the indicator function for the event $E$, \ie, $1_{\{x_i = c\}}$ evaluates to 1 if $x_i = c$ and $0$ otherwise. Note that we have $N_a = C-1$. Furthermore, SNNs are characterized by $C=2$, i.e., by a binary alphabet, and we have $\bms_i(x_i) = x_i \in \{0,1\}$.

Writing $\bms_{i,t} = \bms_i(x_{i,t})$ for simplicity of notation, we have 
\begin{align} \label{eq:DEF}
p_\Theta(\bmx_t | \bmx^{t-1}) \propto 
\exp \bigg\{ \sum_{i \in \set{V}} \Big( \bmtheta_i^\top \bms_{i,t} + \sum_{\delta=1}^\tau \sum_{j \in \set{P}_i} \bms_{j,t-\delta}^\top \bfW_{j,i}^{[\delta]} \bms_{i,t} + \sum_{j \in \set{L}_i} \bms_{j,t}^\top \bfU_{j,i} \bms_{i,t} \Big) \bigg\},
\end{align}
for some memory parameter $\tau > 0$. The model \eqref{eq:DEF} captures causal and lateral connections through the second and third terms of the exponential function, respectively. The model is parameterized by the set $\Theta = \{ \bmtheta, \{\bfW^{[\delta]}\}_{\delta \in \{1, \ldots, \tau\}}, \bfU \}$, whose components are described next.  

First, the vector $\bmtheta_i := [\theta_{i,1}, \ldots, \theta_{i,N_a}]^\top$ contains unit-wise natural parameter for unit $i$, which is collected as $\bmtheta = \{ \bmtheta_i\}_{i \in \set{V}}$. Second, for every pair of units $(j,i) \in \set{E}_{\set{P}}$, the $N_a \times N_a$ matrix $\bfW_{j,i}^{[\delta]}$ describes the causal influence of unit $j$ to unit $i$ after a time lag of $\delta \in \{1,\ldots,\tau\}$ time instants. In particular, the $(a',a)$-th element indicates the causal strength from $s_{j,a'}(x_{j,t-\delta})$ to $s_{i,a}(x_{i,t})$. We use the notations $\bfW^{[\delta]} = \{ \bfW_{j,i}^{[\delta]}\}_{(j,i) \in \set{E}_{\set{P}}}$ and $\bfW = \{\bfW^{[\delta]}\}_{\delta \in \{1,\ldots,\tau\}}$. The structure of the causal graph $\set{G}_{\set{P}}$ defines the position of zeros in $\bfW$: $\bfW_{j,i}^{[\delta]}$ is generally non-zero if $(j,i) \in \set{E}_{\set{P}}$ and is an all-zero matrix otherwise. Note that, in general, we have $\bfW_{j,i}^{[\delta]} \neq \bfW_{i,j}^{[\delta]}$ for $(j,i) \in \set{E}_{\set{P}}$. Third, for every pair of units $(j,i) \in \set{E}_{\set{L}}$, the $N_a \times N_a$ matrix $\bfU_{j,i}$ describes the instantaneous dependence between units $j$ and $i$. In particular, the $(a',a)$-th element indicates the correlation strength between the $a'$-th sufficient statistics $s_{j,a'}(x_{j,t})$ of unit $j$ and the $a$-th sufficient statistics $s_{i,a}(x_{i,t})$ of unit $i$. The structure of the lateral graph $\set{G}_{\set{L}}$ defines the position of zeros in $\bfU = \{\bfU_{j,i}\}_{(j,i) \in \set{E}_{\set{L}}}$: $\bfU_{j,i}$ is generally non-zero if $(j,i) \in \set{E}_{\set{L}}$ and is an all-zero matrix otherwise. Note also that we have $\bfU_{j,i} = \bfU_{i,j}$ for $(j,i) \in \set{E}_{\set{L}}$.

\begin{figure}[t]
  \centering
 \includegraphics[width=0.45\columnwidth]{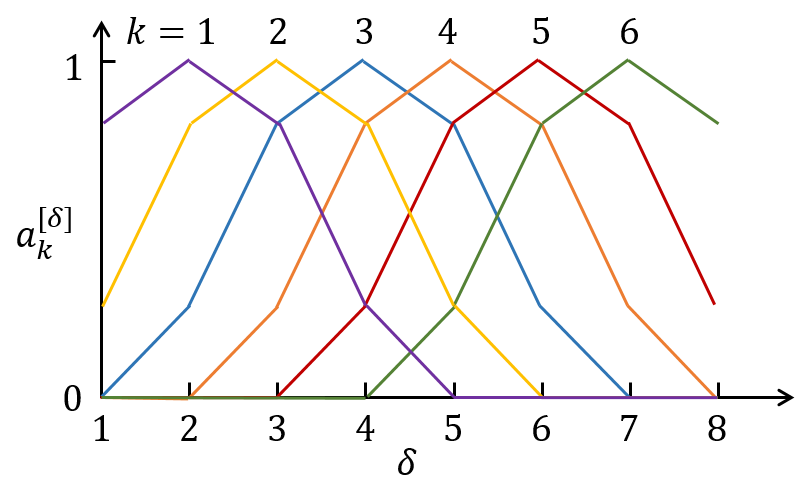}
  \caption{The raised cosine basis functions $a_k^{[\delta]}$, where $k$ is indicated in the figure \cite{pillow08:spatio}. }
    \label{fig:ex_cos}
\end{figure}

Following the common approach in computational neuroscience \cite{pillow08:spatio}, we adopt the parameterization of the causal parameters $\bfW$ as the weighted sum of fixed basis functions with learnable weights. To elaborate, for every $(j,i) \in \set{E}_{\set{P}}$, we write 
\begin{align} \label{eq:basis}
\bfW_{j,i}^{[\delta]} = \sum_{k=1}^K a_k^{[\delta]} \bfV_{j,i,k},
\end{align}
where we have defined $K$ basis functions $a_k^{[\delta]}$, which vary over time $\delta \in \{1,\ldots,\tau\}$ for $k=1,\ldots,K$, and learnable weight matrices $\bfV_{j,i} = [\bfV_{j,i,1}, \ldots, \bfV_{j,i,K}]$ with $\bfV_{j,i,k}$ being a $N_a \times N_a$ matrix, which are collected as $\bfV=\{\bfV_{j,i}\}_{(j,i) \in \set{E}_{\set{P}}}$. We note that a set of basis functions, along with the weight matrices, describes the spatio-temporal receptive field of the neurons \cite{pillow08:spatio}. Examples of basis functions include raised cosines with different synaptic delays \cite{pillow08:spatio, baldi1994delays, taherkhani2015dl}, as illustrated in Fig.~\ref{fig:ex_cos}. 
As a result, the conditional joint distribution in \eqref{eq:DEF} is written as
\begin{align} \label{eq:DEF-basis}
p_\Theta(\bmx_t | \bmx^{t-1}) & \propto 
\exp \Big\{ \sum_{i \in \set{V}} \Big( \bmtheta_i^\top \bms_{i,t} + \sum_{j \in \set{P}_i} \sum_{k=1}^K \Big( \sum_{\delta=1}^\tau a_k^{[\delta]} \bms_{j,t-\delta} \Big)^\top \bfV_{j,i,k} \bms_{i,t} + \sum_{j \in \set{L}_i} \bms_{j,t}^\top \bfU_{j,i} \bms_{i,t} \Big) \Big\},
\end{align}
whose model parameters are defined as $\Theta = \{\bmtheta, \bfV, \bfU\}$. When the number $K$ of basis functions is smaller than $\tau$, the number of parameters is reduced with respect to model \eqref{eq:DEF}.

In \eqref{eq:DEF-basis}, a key role is played by the $k$-th {\em filtered trace} of unit $j$, which we define as the convolution
\begin{align} \label{eq:filter-signal}
\bmalpha_{j,k,t} := \sum_{\delta=0}^{\tau-1} a_k^{[\delta+1]} \bms_{j,t-\delta}. 
\end{align}
This signal is obtained by filtering the signal $\bms_{j,t}$ through a filter with impulse response equal to the basis function $a_k^{[\delta]}$. With the help of this definition, we can re-express the conditional joint distribution in \eqref{eq:DEF-basis} as
\begin{subequations} \label{eq:DEF-filtered}
\begin{align} 
p_\Theta(\bmx_t | \bmx^{t-1}) &= p_\Theta(\bmx_t | \bmalpha_{t-1}) \label{eq:DEF-filtered-a} \\ 
&\propto 
\exp \Big\{ \sum_{i \in \set{V}} \Big( \bmtheta_i + \sum_{j \in \set{P}_i} \sum_{k=1}^K \bfV_{j,i,k}^\top \bmalpha_{j,k,t-1} \Big)^\top \bms_{i,t} + \sum_{(j,i) \in \set{E}_{\set{L}}} \bms_{j,t}^\top \bfU_{j,i} \bms_{i,t} \Big\} \label{eq:DEF-filtered-b}. 
\end{align}
\end{subequations}
In \eqref{eq:DEF-filtered-a}, we have emphasized the dependence of $\bmx_{t}$ on the history $\bmx^{t-1}$ only through the filtered traces $\bmalpha_{t-1} = \{\bmalpha_{j,t-1}\}_{j \in \set{V}}$ with $\bmalpha_{j,t-1} = [\bmalpha_{j,1,t-1}, \ldots, \bmalpha_{j,K,t-1}]$. 

Furthermore, the set of filtered traces $\bmalpha_{j,t-1} = [\bmalpha_{j,1,t-1}, \ldots, \bmalpha_{j,K,t-1}]$ from all units $j$ in $\set{P}_i$ determines the {\em generalized membrane potential} of unit $i$ at time $t$ 
\begin{align} \label{eq:potential}
\bmr_{i,t} = \bmtheta_i + \sum_{j \in \set{P}_i} \sum_{k=1}^K \bfV_{j,i,k}^\top \bmalpha_{j,k,t-1}.
\end{align}
Then, we can rewrite the conditional joint distribution \eqref{eq:DEF-basis} as
\begin{subequations} \label{eq:DEF-potential}
\begin{align} 
p_\Theta(\bmx_t | \bmx^{t-1}) &= p_\Theta(\bmx_t | \bmr_{t} ) \label{eq:DEF-potential-a} \\  
& \propto 
\exp \Big\{ \sum_{i \in \set{V}} \bmr_{i,t}^\top \bms_{i,t} + \sum_{(j,i) \in \set{E}_{\set{L}}} \bms_{j,t}^\top \bfU_{j,i} \bms_{i,t} \Big\} \label{eq:DEF-potential-b}. 
\end{align}
\end{subequations}
In \eqref{eq:DEF-potential-a}, we have emphasized the dependence of $\bmx_t$ on the history $\bmx^{t-1}$ only through the membrane potentials $\bmr_{t} = \{\bmr_{i,t}\}_{i \in \set{V}}$.

\section{Learning} \label{sec:learning}

In this section, we tackle the problem of learning the model parameter $\Theta$ for fixed basis functions under the assumption of fully observed time series. As we will see in Sec.~\ref{sec:numerical}, this set-up is, for instance, relevant to the training of two-layer SNNs and generalizations thereof, in supervised learning problems.

\subsection{Maximum Likelihood Learning via Gradient Descent} \label{sec:mle-learning}

We first aim at deriving gradient-based rules for ML training. To this end, we start by writing the log-likelihood of a given single time-series $\bmx^T$ as 
\begin{align} \label{eq:DEF-ll}
\set{L}_{\bmx^T}(\Theta) := \ln p_\Theta(\bmx^T) = 
\sum_{t=1}^T \ln p_\Theta(\bmx_t | \bmr_t).
\end{align}
The gradient of the log-likelihood function is obtained as 
%\begin{align} \label{eq:DEF-tll-grad}
$\grad_{\Theta} \set{L}_{\bmx^T}(\Theta) = \sum_{t=1}^T \grad_{\Theta}\ln p_\Theta(\bmx_t | \bmr_{t}),$
%\end{align}
where 
\begin{subequations} \label{eq:DEF-ll-grad}
\begin{align} 
\grad_{\bmtheta_i} \ln p_\Theta(\bmx_t | \bmr_t) &= \bms_{i,t} - \mathbb{E}_{\bfx_t \sim p_\Theta(\bmx_t | \bmr_t)} \big[ \bfs_{i,t} | \bmr_t \big], \label{eq:DEF-ll-grad-th} \\
\grad_{\bfV_{j,i}} \ln p_\Theta(\bmx_t | \bmr_t) &= \bmalpha_{j,t-1} \big( \bms_{i,t} - \mathbb{E}_{\bfx_t \sim p_\Theta(\bmx_t | \bmr_t)} \big[ \bfs_{i,t} | \bmr_t \big] \big)^\top, \label{eq:DEF-ll-grad-V} \\
\grad_{\bfU_{j,i}} \ln p_\Theta(\bmx_t | \bmr_t) &=  \bms_{j,t}\bms_{i,t}^\top - \mathbb{E}_{\bfx_t \sim p_\Theta(\bmx_t | \bmr_t)} \big[ \bfs_{j,t}\bfs_{i,t}^\top | \bmr_t \big]. \label{eq:DEF-ll-grad-U} 
\end{align}
\end{subequations}
The gradients \eqref{eq:DEF-ll-grad} have the structure, typical for exponential family models, that presents the difference between the empirical average of the relevant sufficient statistics given the actual realization of the time series $\bmx^T$ and the corresponding ensemble average under the model distribution \eqref{eq:DEF-potential}. In detail, the gradient in \eqref{eq:DEF-ll-grad} has two components: {\em (i)} the {\em positive} component points in a direction of the natural parameter space that maximizes the unnormalized log-distribution, or energy, \ie, the logarithm of \eqref{eq:DEF-potential-b}, for a given data $\bmx^T$, thus maximizing the fitness of the model to the observed data $\bmx^T$; and {\em (ii)} the {\em negative} component points a direction that minimizes the partition function, \ie, the corresponding normalization term, hence minimizing the fitness of the model to the unobserved configurations (see, \eg, \cite{simeone2018brief}).

\begin{algorithm}[t]
\setstretch{0.975}
%\SetAlgoNoLine
\LinesNumbered
%\smallskip
\KwIn{Causal graph $\set{G}_{\set{P}}$, lateral graph $\set{G}_{\set{L}}$, training set $\set{D}$ of time-series $\bmx^T$, and learning rate $\eta$}
%\vspace{-0.1cm}
\KwOut{Learned model parameters $\Theta$}
\vspace{0.1cm}
\hrule
\vspace{0.1cm}
{\bf initialize} parameters $\Theta$ \\
%\vspace{-0.2cm}
\Repeat{{\em convergence of $\Theta$}}{
%\vspace{-0.2cm}
draw an example $\bmx^T$ from the training set $\set{D}$ \\
%\vspace{-0.2cm}
\vspace{0.05cm}
\lFor{{\em each unit $i \in \set{V}$}}{compute the gradients $\grad_{\Theta}\set{L}_{\bmx^T}(\Theta)$ in \eqref{eq:DEF-ll-grad} based on $\bmr_{\set{C}_i \cup \{i\},t}$ and $\Theta_i \cup \bfU_{\set{C}_i}$ (see Fig.~\ref{fig:grad_info})}
\vspace{0.05cm}
update model parameters
$$\Theta \leftarrow \Theta + \eta \grad_{\Theta} \set{L}_{\bmx^T}(\Theta)$$
}
\caption{Maximum Likelihood Learning via Gradient Descent}
\label{alg:dec_mle}
\end{algorithm}

\begin{figure}[t]
  \centering
  \hspace{-0.2cm}
  \subfigure[]
  {
    \includegraphics[height=0.185\columnwidth]{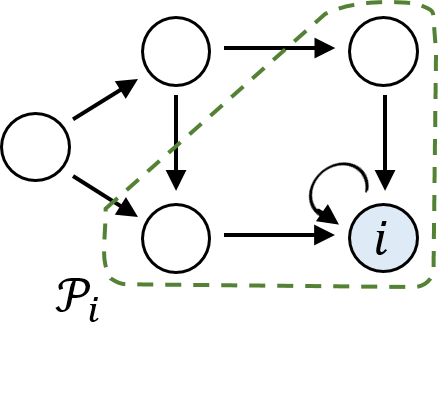}
    \label{fig:info_G_P}
  } 
  \hspace{-0.1cm}
  \subfigure[]
  {
    \includegraphics[height=0.183\columnwidth]{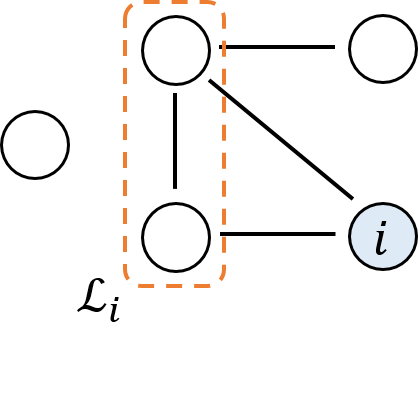}
    \label{fig:info_G_L}
  } 
  \hspace{-0.1cm}
  \subfigure[]
  {
    \includegraphics[height=0.185\columnwidth]{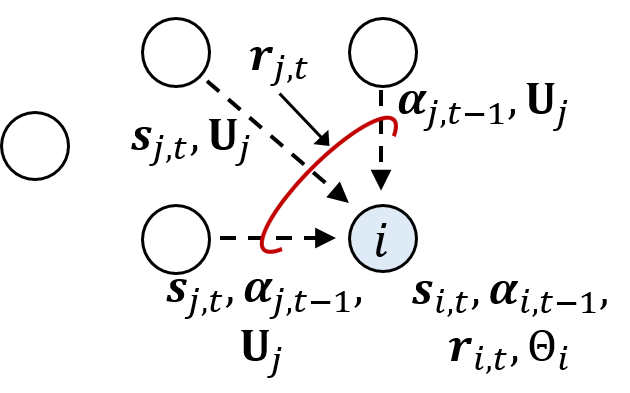}
    \label{fig:info_local}
  } 
  \hspace{-0.25cm}
  \subfigure[]
  {
    \includegraphics[height=0.185\columnwidth]{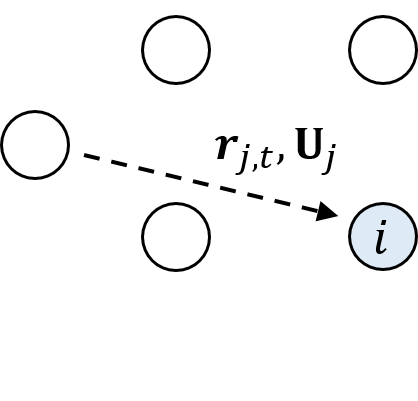}
    \label{fig:info_nonlocal}
  }   
  \hspace{-0.15cm}
  \caption{Illustration of the information required to compute the gradients \eqref{eq:DEF-ll-grad} for unit $i$: (a) directed graph $\set{G}_{\set{P}}$; (b) undirected graph $\set{G}_{\set{L}}$; (c) local and (d) non-local information necessary for learning. }
  \label{fig:grad_info}
  %\vspace{-0.3cm}
\end{figure}

A stochastic gradient-based optimizer iteratively updates the parameter $\Theta$ based on the rule summarized in Algorithm~\ref{alg:dec_mle}, where the learning rate $\eta$ is assumed to be fixed here for simplicity of notation. With this rule, each unit $i \in \set{V}$ updates its own local parameters $\Theta_i = \{ \bmtheta_i, \bfV_i, \bfU_i \}$, where we use the notations $\bfV_i = \{ \bfV_{j,i}\}_{j \in \set{P}_i}$ and $\bfU_i = \{\bfU_{j,i}\}_{j \in \set{L}_i}$. We note that intersections of the sets of local parameters $\Theta_i$ are non-empty in the presence of the lateral connections among units, and the lateral parameters $\bfU$ can be updated at all the units connected to edges $\set{E}_{\set{L}}$ in a consistent manner. Details of the distributed computation of the gradients are provided next (see also Fig.~\ref{fig:grad_info}).

\smallskip
\noindent {\bf Distributed computation of the gradients.} We now discuss the information required to compute the gradients \eqref{eq:DEF-ll-grad} for each unit $i$ in order to enable a distributed implementation of Algorithm~\ref{alg:dec_mle}. We define as {\em local} to neuron $i$ all quantities that concern units that have either a directed edge to neuron $i$ in $\set{G}_{\set{P}}$ or an undirected edge with it in $\set{G}_{\set{L}}$. The positive components of all the gradients \eqref{eq:DEF-ll-grad} require the local statistics $\bms_{i,t} = \bms_i(x_{i,t})$ and the local filtered traces $\bmalpha_{j,t-1}$ for all parents $j \in \set{P}_i$. Furthermore, they also require the current local sufficient statistics $\bms_{j,t}$ for all units $j \in \set{L}_i$ connected by lateral edge. 
In contrast, the negative components entail the computation of the average of the local sufficient statistics $\bfs_{i,t}$ and of the products $\bfs_{j,t} \bfs_{i,t}^\top$ for $j \in \set{L}_i$ over their marginal distributions, which is further discussed next. 

The marginal distribution of the local sufficient statistics $\bfs_{i,t}$ is obtained from $p_\Theta(x_{i,t} | \bmr_t)$
\begin{align} \label{eq:DEF-marginal-node}
p_\Theta(x_{i,t}| \bmr_t) = \sum_{\bmx_{\set{C}_i,t}} p_{ \Theta_i \cup \bfU_{\set{C}_i} }(x_{i,t}, \bmx_{\set{C}_i,t} | \bmr_{\set{C}_i \cup \{i\},t}),
\end{align}
where we recall that $\set{C}_i$  is the subset of units that are reachable from unit $i$ via lateral connections. As indicated by the notation in \eqref{eq:DEF-marginal-node}, the joint distribution on the right-hand side can be obtained from \eqref{eq:DEF-potential-b} by including in the sums at the exponent only the terms that depend on $\bms_{i,t}$ and $\bms_{\set{C}_i,t}$. Calculation of \eqref{eq:DEF-marginal-node} hence requires knowledge of the membrane potentials from the units in $\set{C}_i \cup \{i\}$, and it depends on the local parameters $\Theta_i$ and lateral parameters $\bfU_{\set{C}_i} = \{ \bfU_{j} \}_{j \in \set{C}_i}$ of all units $j$ in the set $\set{C}_i$. The marginal distribution of the products $\bfs_{j,t} \bfs_{i,t}^\top$ for $(j,i) \in \set{E}_{\set{L}}$ is similarly computed as 
\begin{align} \label{eq:DEF-marginal-edge}
p_\Theta(x_{j,t},x_{i,t} | \bmr_t) = \sum_{\bmx_{\set{C}_i \setminus \{j\},t}} p_{\Theta_i \cup \bfU_{\set{C}_i}}(x_{j,t},x_{i,t}, \bmx_{\set{C}_i \setminus \{j\},t} | \bmr_{\set{C}_i \cup \{i\},t}).
\end{align}

In summary, as illustrated in Fig.~\ref{fig:grad_info}, the computation of the gradients \eqref{eq:DEF-ll-grad} for unit $i$ requires knowledge of the local sufficient statistics $\bms_{i,t}$ and $\bms_{j,t}$ for all $j \in \set{L}_i$, of the local filtered traces $\bmalpha_{i,t-1}$ and $\bmalpha_{j,t-1}$ for all $j \in \set{P}_i$, as well as the local membrane potentials $\bmr_{j,t}$ and the current lateral parameters $\bfU_j$ for all $j \in (\set{L}_i \cup \set{P}_i) \cap \set{C}_i$. It also requires the acquisition of the membrane potentials $\bmr_{j,t}$ and of the current lateral parameters $\bfU_j$ for all units $j$ in the set $\set{C}_i$. This calls for non-local signaling if there are units in $\set{C}_i$ that are not in the sets $\set{L}_i$ or $\set{P}_i$.

The complexity of the computation of the second terms in \eqref{eq:DEF-ll-grad} depends exponentially on the size of the set $\set{C}_i$. In particular, if there are no lateral connections, \ie, $\set{G}_{\set{L}}$ is an empty graph, the marginalizations \eqref{eq:DEF-marginal-node}-\eqref{eq:DEF-marginal-edge} are not required since the set $\set{C}_i$ of every unit $i$ is empty, and the second terms can be obtained in closed form. When the sets $\set{C}_i$ are too large to enable computation of the sums in \eqref{eq:DEF-marginal-node}-\eqref{eq:DEF-marginal-edge}, Gibbs sampling, or approximations such as Contrastive Divergence, can be used to enable the expectation \cite{hinton02:cd}.

\subsection{Bayesian Learning} \label{sec:langevin-learning}

As a potentially more powerful alternative to ML learning, Bayesian methods aim at computing the posterior distribution of the parameter vector $\Theta$ under the assumption of a given prior $p(\Theta)$. Bayesian learning can capture parameter uncertainty and is able to naturally avoid overfitting. While exact Bayesian learning is prohibitively complex, a gradient-based method has been recently proposed that combines Robbins-Monro type stochastic approximation methods \cite{robbins51:stochastic} with Langevin dynamics \cite{neal11:mcmc}. The method, called the {\em stochastic gradient Langevin dynamics} \cite{welling11:langevin, patterson13:langevin, sato14:langevin,kappel15:synaptic}, updates the parameter $\Theta$ based on the rule summarized in Algorithm~\ref{alg:langevin}, where $\eta$ is the learning rate and ${\bm \nu}$ denotes i.i.d. Gaussian noise with zero mean and variance $1$. The update rule in Algorithm~\ref{alg:langevin} has the property that the distribution of the parameter vector $\Theta$ at equilibrium matches the Bayesian posterior as $\eta \rightarrow 0$.

\begin{algorithm}[t]
\setstretch{0.975}
%\SetAlgoNoLine
\LinesNumbered
%\smallskip
\KwIn{Causal graph $\set{G}_{\set{P}}$, lateral graph $\set{G}_{\set{L}}$, training set $\set{D}$ of time-series $\bmx^T$, prior of the parameters $p(\Theta)$, and learning rate $\eta$}
%\vspace{-0.1cm}
\KwOut{Learned model parameters $\Theta$}
\vspace{0.1cm}
\hrule
\vspace{0.1cm}
{\bf initialize} parameters $\Theta$ \\
%\vspace{-0.2cm}
\Repeat{{\em convergence of $\Theta$}}{
%\vspace{-0.2cm}
draw an example $\bmx^T$ from the training set $\set{D}$ \\
%\vspace{-0.2cm}
\vspace{0.05cm}
\lFor{{\em each unit $i \in \set{V}$}}{compute the gradients $\grad_{\Theta}\set{L}_{\bmx^T}(\Theta)$ in \eqref{eq:DEF-ll-grad} based on $\bmr_{\set{C}_i \cup \{i\},t}$ and $\Theta_i \cup \bfU_{\set{C}_i}$ (as in Algorithm~\ref{alg:dec_mle})}
\vspace{0.05cm}
update model parameters
$$\Theta \leftarrow \Theta + \eta \bigg( \grad_\Theta \ln p(\Theta) + |\set{D}| \cdot  \grad_{\Theta} \set{L}_{\bmx^T}(\Theta) \bigg) + \sqrt{2 \eta} {\bm \nu}$$
}
\caption{Bayesian Learning}
\label{alg:langevin}
\end{algorithm}

\section{Results and Discussion} \label{sec:numerical}

\begin{figure}[ht]
  \centering
  \includegraphics[width=0.57\columnwidth]{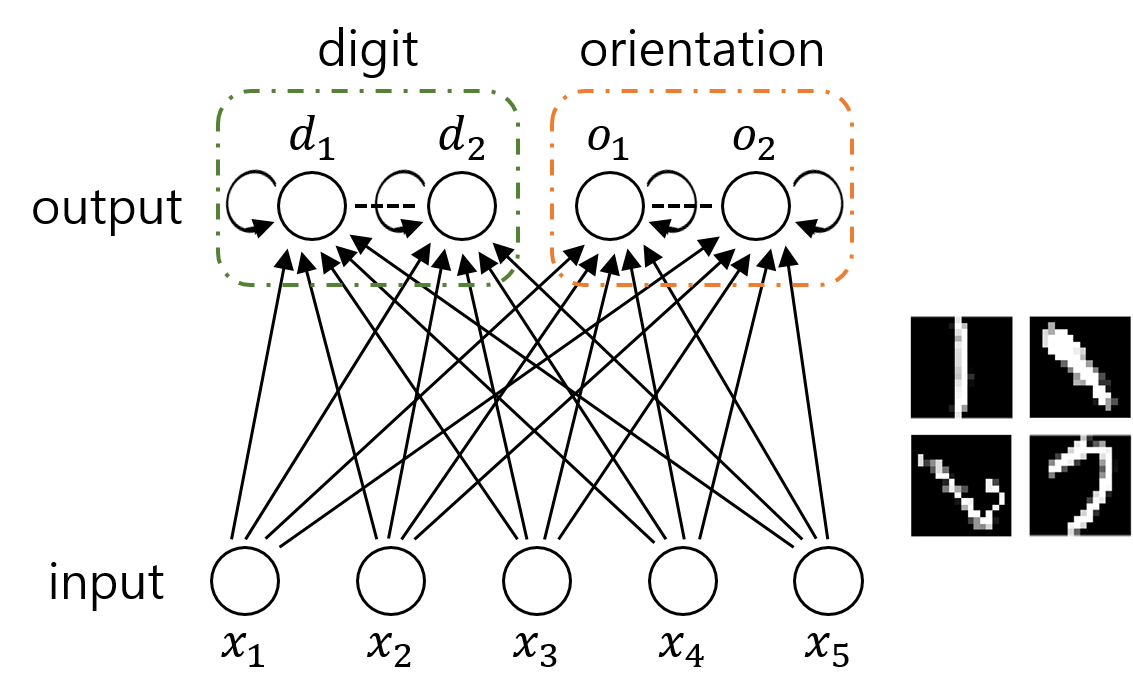}
  \caption{Hybrid directed and undirected graphical architecture of a two-layer SNN used for supervised learning: each input neuron $x_i$ corresponds to a pixel of the input image; each output neuron $d_i$ encodes the digit index; and each output neuron $o_i$ encodes the orientation index. The directed graph has causal connections from input neurons to output neurons and self-loops at the output neurons; and the undirected graph has lateral connections only between output neurons in the same subset.}
  \label{fig:supervised_SNN}
%  \vspace{-0.3cm}
\end{figure}

In this section, we consider an application of the methods developed in this paper to a {\em multi-task} supervised learning problem based on a two-layer SNN. The problem consists of the two tasks of classifying a handwritten number and of detecting a possible rotation of the image of the handwritten digit (see Fig.~\ref{fig:supervised_SNN}). As illustrated in Fig.~\ref{fig:supervised_SNN}, the set of neurons $\set{V}$ consists of two layers: the spike trains associated with the neurons in the first layer are determined by the input image, while the spike trains corresponding to the second layer are determined by the correct labels for the indices of number and orientation of the image. Specifically, we have one input neuron for every pixel of the image, and the output layer contains two subsets of neurons: one subset of neurons with one neuron for every possible digit index, and another subset with one neuron for each possible orientation, i.e., vertical (unrotated) or rotated. The two-layer SNN can be represented as a dynamic exponential family, in which there are causal connections only from input neurons to output neurons, with self-loops at the output neurons, as well as lateral connections between output neurons in the same subset in order to capture the correlations of the labels in the same task. 

The training dataset is generated by selecting $500$ images of each digit ``$1$'' and ``$7$'' from the USPS handwritten digit dataset \cite{hull1994database}. All $16 \times 16$ images are included as they are, that is, unrotated, and they are also included upon a rotation by a randomly selected degree. The test set is similarly obtained by using $125$ examples from the USPS dataset. As a result, we have $256$ input neurons, with one per pixel of the input image. Each gray scale pixel is converted to an input spike train by generating an i.i.d. Bernoulli vector of $T$ samples with spike probability proportional to the pixel intensity and limited between $0$ and $0.5$. The digit labels $\{1,7\}$ are encoded by the first subset of two output neurons, so that the desired output of the neuron corresponding to the correct label index is one spike emission every three non-spike samples, while the other neuron is silent. The same is done for the orientation labels $\{\mathtt{v}, \mathtt{r}\}$, using the second subset of two output neurons, where the label $\mathtt{v}$ indicates the original unrotated image, and the label $\mathtt{r}$ indicates the rotated image. In order to test the classification accuracy, we assume standard rate decoding, which selects the output neuron, and hence the corresponding indices, with the largest number of output spikes. Finally, in order to investigate the impact of lateral connections in our model, we also consider a two-layer SNN in which lateral connections are not allowed. 

\begin{figure}[t]
  \centering
  \includegraphics[width=0.57\columnwidth]{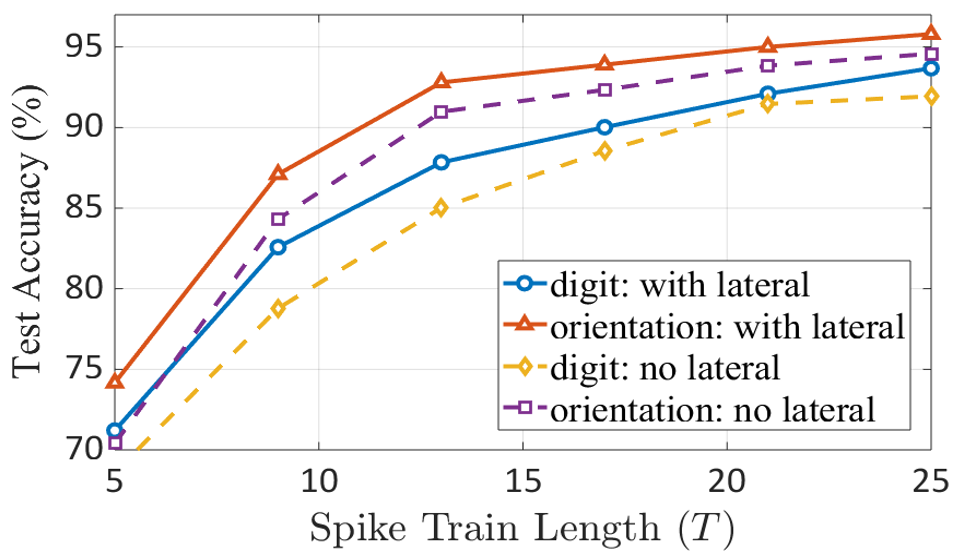}
  \caption{ML learning: test accuracy for the two tasks of digit and orientation classification versus the duration $T$ of the time series when $K=2$.}
  \label{fig:two_rnd_T_K2}
  %\vspace{-0.15cm}
\end{figure}

First, we train both models with the ML learning rule in Algorithm~\ref{alg:dec_mle} for $800$ epochs with constant learning rate $\eta = 0.05$, based on $20$ trials with different random seeds. The model parameters are randomly initialized with uniform distribution in range $[-1,1]$, while the lateral parameters are randomly initialized in range $[-2,2]$. We assume the $K$ raised cosine basis functions introduced in \cite{pillow08:spatio,bagheri18:snn_first} in \eqref{eq:basis}. Fig.~\ref{fig:two_rnd_T_K2} demonstrates the classification accuracy in the test set versus the length $T$ of time series with $K=2$. From the figure, the task of classifying orientation is seen to be easier than digit identification. Furthermore, we observe better accuracies in both tasks when we train the model with lateral connections between output neurons. This implies that Algorithm~\ref{alg:dec_mle} can efficiently learn how to make use of the instantaneous correlations among output neurons participating in the same task.

\begin{figure}[t]
%  \hspace{-0.2cm}
  \centering
  \subfigure[]
  {
	\centering
\includegraphics[height=0.3\columnwidth]{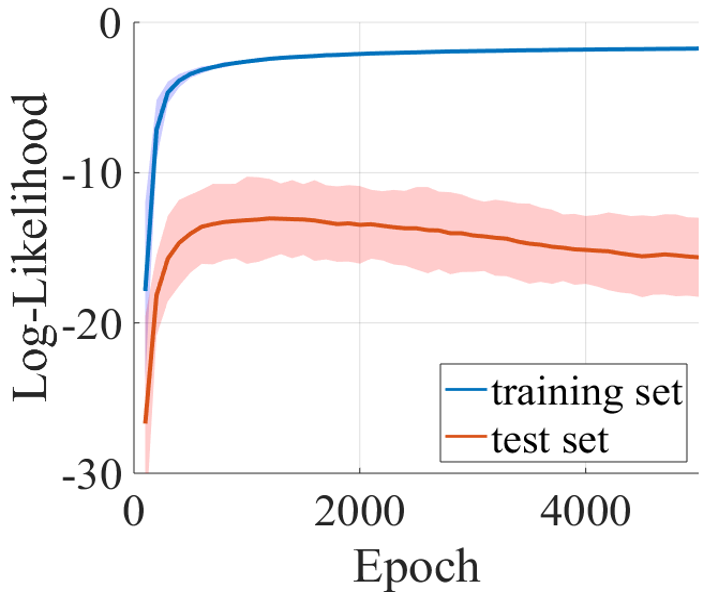}
    \label{fig:langevin_uniform_ll}
  } 
  \hspace{0.2cm}
  \subfigure[]
  {
  \centering
 \includegraphics[height=0.28\columnwidth]{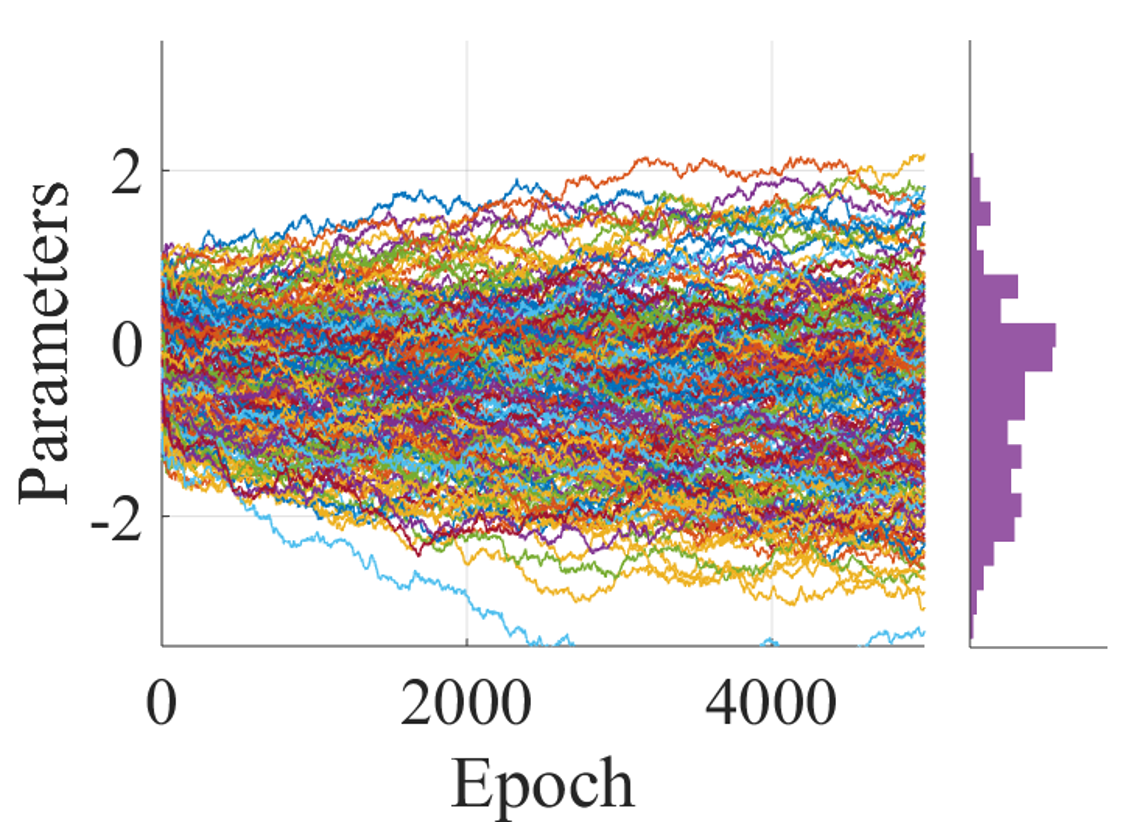} 
 	\label{fig:langevin_uniform_param_histo}
  }  
%  \hspace{-0.2cm}
  \subfigure[]
  {
	\centering
\includegraphics[height=0.3\columnwidth]{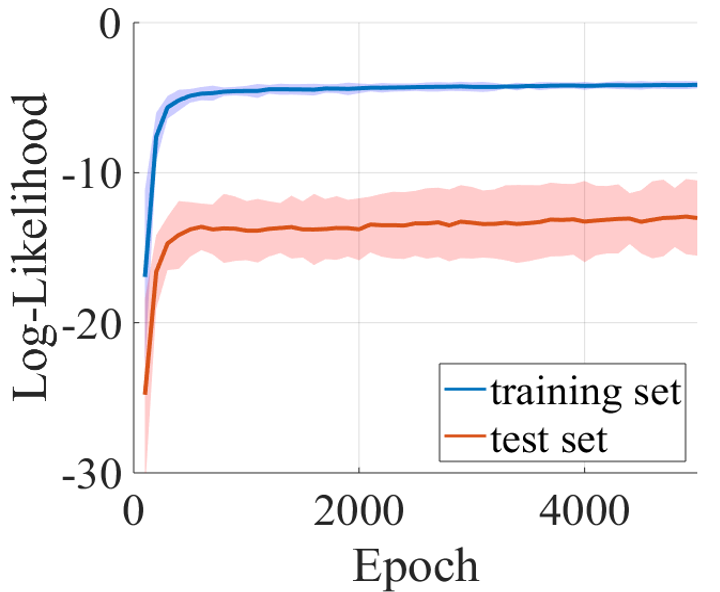}
    \label{fig:langevin_gmm_ll}
  } 
  \hspace{0.2cm}
  \subfigure[]
  {
  \centering
 \includegraphics[height=0.28\columnwidth]{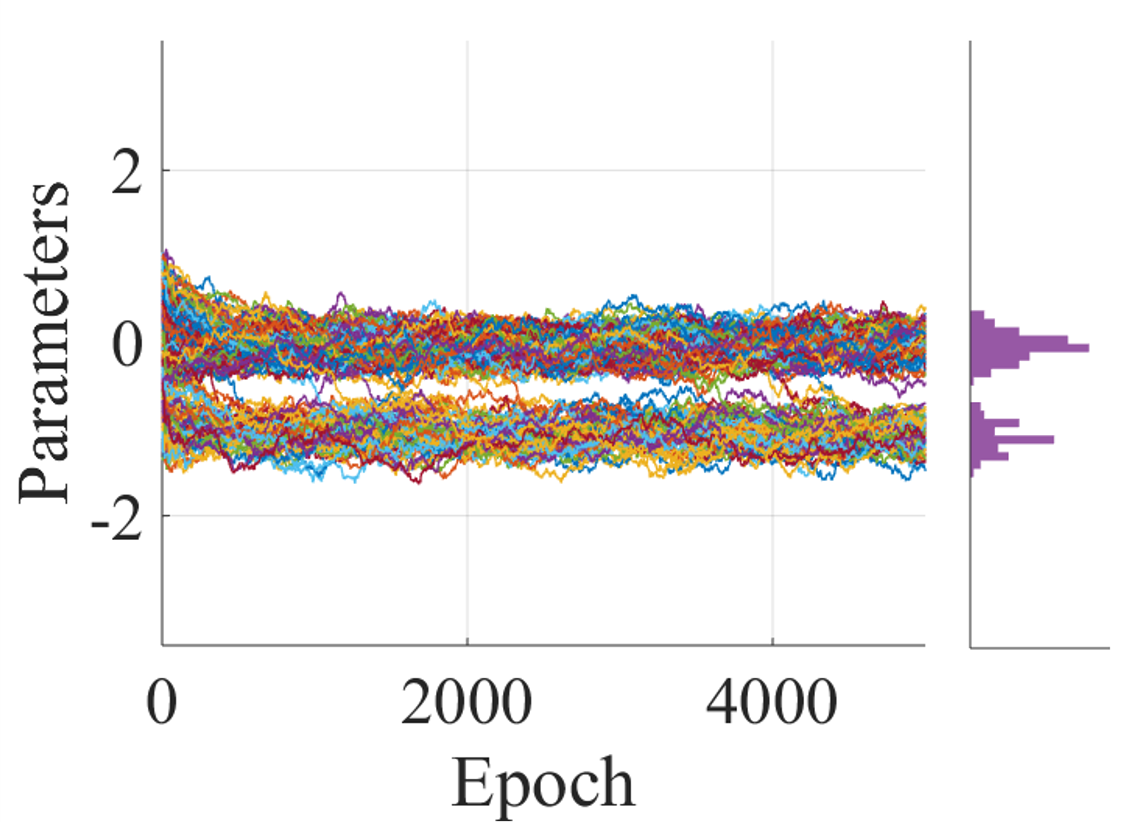} 
 	\label{fig:langevin_gmm_param_histo}
  }  
  \caption{Bayesian learning with different priors: (a)-(b) with uniform prior and (c)-(d) with bimodal prior; figures (a) and (c) show the log-likelihood on the training and test sets as the function of the learning epochs, where the shaded area indicates three standard deviations of the mean based on $20$ trials; and figures (b) and (d) plot the evolution of $80$ randomly selected causal parameters throughout the epochs (left) and the histogram at the end of the learning epoch (right).}
  \label{fig:langevin}
\end{figure}

We then consider the Bayesian learning rule in Algorithm~\ref{alg:langevin}, and study the impact of the prior $p(\Theta)$ for the model parameters on overfitting. To this end, we select $10$ and $50$ samples of each digit and orientation class for training and testing, respectively. After training the model for a number of epochs with constant learning rate $\eta = 0.000625$, we evaluate the learned model by measuring the log-likelihood of the desired output spikes for the correct label given the input images in both training and test set. For the experiment in Fig.~\ref{fig:langevin_uniform_ll}-\ref{fig:langevin_uniform_param_histo}, a uniform prior over the causal parameters $\bfW$ is assumed; while we choose a mixture of two Gaussians with means $0$ and $-1.0$ and identical standard deviations $0.15$ as a prior for the results in Fig.~\ref{fig:langevin_gmm_ll}-\ref{fig:langevin_gmm_param_histo}, following the approach in \cite{kappel15:synaptic}. Fig.~\ref{fig:langevin_uniform_ll} shows that a uniform prior can lead to overfitting, as also suggested by the larger variance of the sampled weights plotted in Fig.~\ref{fig:langevin_uniform_param_histo}. In contrast, a bimodal prior yields good generalization performance due to its capacity to act as a regularizer by keeping a fraction of the weights close to zero as seen in Fig.~\ref{fig:langevin_gmm_param_histo}.

\bibliographystyle{IEEEtran}
\bibliography{ref}

\end{document}